\documentclass[sigconf,screen]{acmart}

\usepackage{latexsym}
\usepackage[T1]{fontenc}
\usepackage[utf8]{inputenc}
\usepackage{listings}
\usepackage{color}
\usepackage{balance} 
\usepackage{booktabs}

\definecolor{dkgreen}{rgb}{0,0.6,0}
\definecolor{gray}{rgb}{0.5,0.5,0.5}
\definecolor{mauve}{rgb}{0.58,0,0.82}
\definecolor{backcolour}{rgb}{0.95,0.95,0.92}

\copyrightyear{2023}
\acmYear{2023}
\setcopyright{othergov}
\acmConference[CODS-COMAD 2023]{6th Joint International Conference on Data Science \& Management of Data (10th ACM IKDD CODS and 28th COMAD)}{January 4--7, 2023}{Mumbai, India}
\acmBooktitle{6th Joint International Conference on Data Science \& Management of Data (10th ACM IKDD CODS and 28th COMAD) (CODS-COMAD 2023), January 4--7, 2023, Mumbai, India}
\acmPrice{15.00}
\acmDOI{10.1145/3570991.3571068}
\acmISBN{978-1-4503-9797-1/23/01}
\usepackage{fancyvrb}
\usepackage{pifont}

\begin{document}

\title{UDAAN - Machine Learning based Post-Editing tool for Document Translation }
\author{Ayush Maheshwari, Ajay Ravindran, Venkatapathy Subramanian, Ganesh Ramakrishnan }
\email{{ayusham, ajayr, venkatapathy, ganesh}@cse.iitb.ac.in}
\affiliation{%
  \institution{Indian Institute of Technology Bombay}
  \country{India}
  }

\renewcommand{\shortauthors}{A. Maheshwari et al.}
\begin{abstract}
We introduce UDAAN, an open-source post-editing tool that can reduce manual editing efforts to quickly produce publishable-standard documents in several Indic languages. 
UDAAN has an end-to-end Machine Translation (MT) plus post-editing pipeline wherein users can upload a document to obtain raw MT output. Further, users can edit the raw translations using our tool. UDAAN offers several advantages: i. Domain-aware, vocabulary-based lexical constrained MT. ii. source-target and target-target lexicon suggestions for users. Replacements are based on the source and target texts’ lexicon alignment. iii. Translation suggestions are based on logs created during user interaction. iv. Source-target sentence alignment visualisation that reduces the cognitive load of users during editing. v. Translated outputs from our tool are available in multiple formats: docs, latex, and PDF.
We also provide the facility to use around 100 in-domain dictionaries for lexicon-aware machine translation.
Although we limit our experiments to English-to-Hindi translation, our tool is independent of the source and target languages. Experimental results based on the usage of the tools and users' feedback show that our tool speeds up the translation time by approximately a factor of three compared to the baseline method of translating documents from scratch. Our tool  is available for both \href{https://drive.google.com/drive/folders/1nE2-a71IeubdFPOceV9MX7rNB96Q2Skl?usp=sharing}{Windows} and \href{https://drive.google.com/drive/folders/1wgJgJgdPdbUEyFI_z9izXw1tq80p7gKR}{Linux platforms}.
The tool is open-source under MIT license, and the source code can be accessed from our website, \url{https://www.udaanproject.org}.
Demonstration and tutorial videos for various features of our tool can be accessed \href{https://www.youtube.com/channel/UClfK7iC8J7b22bj3GwAUaCw}{here}. Our MT pipeline can be accessed at \url{https://udaaniitb.aicte-india.org/udaan/translate/}.
\end{abstract}


\maketitle


\section{Introduction}
Recently, neural-based machine translation (NMT) systems have led to tremendous improvement in the machine translation performance~\cite{barrault-etal-2019-findings}. However, machine translation (MT) quality is still far from the quality of human translation~\cite{freitag2021experts}. In practice, human translators perform post-editing (PE) on top of the output produced by NMT systems with the help of computer-aided translation (CAT) tools. 

\begin{table}[!ht]
\centering
\begin{tabular}{p{0.08\linewidth} | p{0.25\linewidth}| p{0.60\linewidth}}
\toprule
Term & Meaning                    & \multicolumn{1}{c}{Description}                                                                     \\ \midrule
MT           & Machine Translation        & System used to automatically translate text                                                         \\
CAT          & Computer-aided Translation & Software used to assist a human translator in the translation process.                              \\
TM           & Translation Memory         & Database of stored segments of already translated sentences. \\
PE           & Post-editing               & Process adopted by human to refine translation produced by MT.                                      \\ \bottomrule
\end{tabular}%
\caption{Key concepts used in the paper.}
\label{tab:abbr}
\end{table}

\begin{figure*}[t]
\centering
    \includegraphics[width=0.9\linewidth]{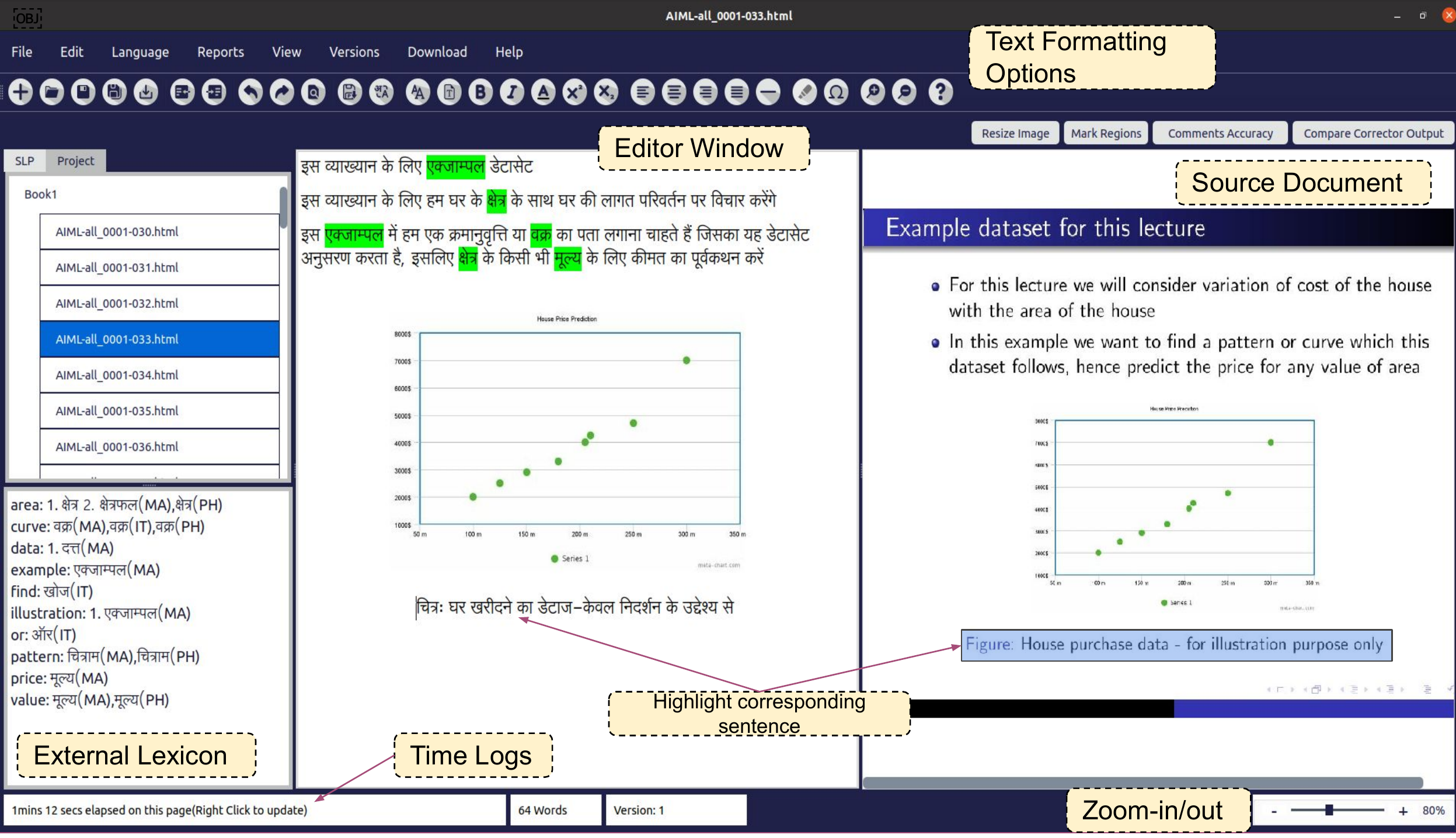}
    \caption{Screenshot of the UDAAN tool with highlights about its key features. 
    }
    \label{fig:tool-image}
\end{figure*}

Existing CAT tools rely on translation memory (TM) so that translated pairs can be reused later, like a find-and-replace function but between two languages. Additionally, TMs are integrated with additional domain-specific terminologies in a format similar to bilingual glossaries. However, using TMs results in over-recycling of sentences (or their parts)  \cite{bedard2000memoire}. Translators may use TM deliberately to reduce variance in the translation resulting in poor translation quality and inconsistency \cite{moorkens2014virtuous, doherty2016translations}. We circumvent this problem by allowing the user to ingest domain-specific lexicons during the stage of generating raw MT. In addition to the interactive translation suggestions, we reduce post-editing efforts by suggesting the users (optionally) to apply edits made on the current page across all pages, with the additional visual context for correct replacement decisions.

Current tools merely ingest sentences without providing any support for PDF or image files. We provide Optical Character Recognition (OCR) support for non-parse-able document formats, such as PDF and image files as well. 
 In this paper, we introduce UDAAN, an end-to-end document translation pipeline with an efficient PE tool that combines both the benefits of TM and MT. We integrate both OCR and MT pipeline, thus allowing users to upload PDF or image for translation, reducing the additional overhead of OCR conversion\footnote{We can use our tool for OCR post-editing as well \cite{ocr-sanskrit}.}. To mitigate the overall load for translators, we introduce following features in our tool:

 \noindent \textbf{Interactive Translation}: The UDAAN tool provisions for interactive translation and post-editing. By learning from the edit patterns of a user, we provide for either ``global replacements'' or ``local suggestions for replacements'' based on similar patterns. This reduces editing time and ensures consistency in the edited output. 
    
\noindent\textbf{Lexically Constrained Translation}: Often, a model trained on a particular domain does not reliably work on another domain sentence. We circumvent this problem by training a lexically constrained translation model\cite{chen2021lexically} that accepts a source-target lexicon and a source sentence as input and produces a target translation constrained towards the input lexicon. The target lexicon is ingested in a probabilistic manner instead of hard replacement.  This prevents the requirement to train domain-specific models while inducing domain vocabulary in the target translation thus producing domain-aware translation and significantly reducing the post-editing effort. Our online interface \footnote{https://udaaniitb.aicte-india.org/udaan/translate} provides the facility to choose from > 100 dictionaries.

 \noindent \textbf{Lexicon suggestions}: Besides the machine-generated translation for the entire source document, domain-specific technical dictionaries (source-target) are presented to guide the user's edits. The tool allows users to track modifications based on global replacements or technical dictionary suggestions. This tracking is maintained by highlighting the words using a variety of colours.

\noindent \textbf{Sentence alignment}: To reduce the cognitive load on the users, we highlight sentences in the source as well as the translated documents, as shown in Figure \ref{fig:tool-image}. Source and target sentences are highlighted on mouse hover over the corresponding sentences in either window pane(source/target contents). Alignment is preserved by tracking unique ids in the source and target sentences. In the case of image files, the unique IDs are also mapped to their corresponding bounding boxes detected by the OCR engine. 


\noindent \textbf{Rich Text Formatting}: Users can perform rich text editing similar to word processor software. Our tool allows adding images, equations, tables \emph{etc.}, and allowing cropping and resizing of images. We can export the final post-edited document in various popular formats such as docx, pdf, latex, \emph{etc.}

\noindent \textbf{Offline working and sync facility}: Often, translators prefer working offline using their desktops. The tool can be used in a completely offline mode on a desktop (Windows or Linux). Continuous access to the internet is not required.
Additionally, our tool offers version control and synchronisation using Git for continuous backup and seamless collaboration.
    
\noindent \textbf{Multiple roles}: Our tool can be accessed with different roles such as \textbf{Correctors}- who edits a document, \textbf{Verifiers}- who validate the output from Correctors, and \textbf{Proofreaders}- who validate the final translated document.

\noindent Most of our target languages are of the \textit{Indic} language family, which is typed in the Devanagari script. In addition, the tool allows users to enter text through SLP1\footnote{SLP1 is an ASCII transliteration scheme for the Sanskrit language from and to the Devanagari script.} format which is found to be more convenient by several users. 
\begin{figure}[!t]
    \centering
    \includegraphics[width=1\linewidth]{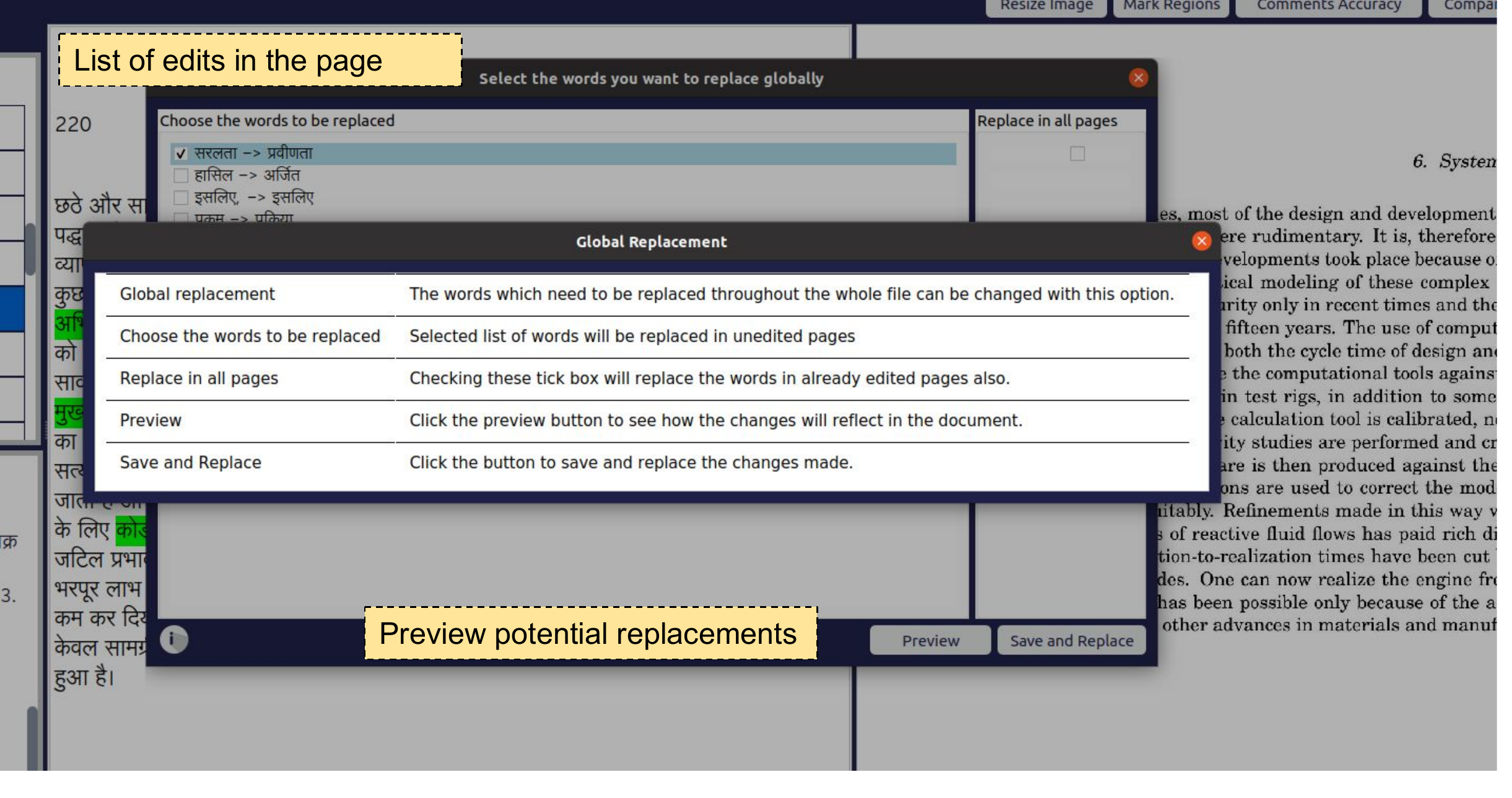}
    \caption{Dialog box listing all the post-edits on the current page. The user is shown a list of edits and the option to replace them in (i) unedited pages and (ii) all the pages included already edited. Users can preview the potential replacements by clicking the Preview button.}
    \label{fig:replacement}
\end{figure}
\section{Related Work}
Professional translators often use CAT (computer-aided translation) tools \cite{van2015recommendations} for localisation tasks. Due to drastic improvements in NMT models, MT has improved substantially and is now actively integrated into the CAT tools \cite{lee2021intellicat, federico2014matecat, herbig2020mmpe, santy2019inmt}. Recent studies demonstrate that post-editing MT output reduces translation errors and improves translation productivity\cite{toral2018post}. Additionally, these tools provide features like translation memory with quality estimation and concordance functionality \cite{federico2014matecat}, alignment between source and MT \cite{schwartz2015effects}, translation suggestion \cite{lee2021intellicat} auto-completion assistance or intelligibility assessments \cite{vandeghinste2019improving}. However, these tools require source documents in a parse-able format such as XML, JSON, doc, \emph{etc.} and work properly for post-editing short documents. To the best of our knowledge, CAT tools are neither designed for long-document translations nor accept source documents in PDF or other difficult-to-parse formats.

\section{System Description}


UDAAN is an offline interactive interface for post-editing MT outputs. It provides the facility to upload a PDF and retrieve the translated document. Along with NMT, OCR is integrated into the tool pipeline. Our tool can be used for post-editing OCR outputs as well. To enable active collaboration among translators and continuous backup of the work, we upload the document translation files over GitHub\footnote{GitHub is an online collaborative software development and version control platform.}. However, we can upload project files over any remote version control system.

Users can upload a PDF document  \href{https://udaaniitb.aicte-india.org/udaan/}{here} and download translated output as a zip file. When the zip opened with our tool, the source document and intial MT output are displayed side-by-side: the uploaded original document (PDF, image, \emph{etc.}) on the right and machine-translated document on the left (see Figure \ref{fig:tool-image}).  The MT output is split into multiple pages retaining the \textit{formatting style} of the original document ({\em e.g.}, alignment). Non-textual parts of a document, such as images, tables and equations are also embedded in the MT document. The tool provides the facility to mark and insert additional images, equations and tables in the PE window. 

\subsection{Optical Character Recognition}
Several existing PE tools assume that input documents are available in clean and parseable formats (e.g., xml, json, docx, xls). However, users often come across documents available in PDF, and image formats ({\em e.g.}, receipts, journal papers, government reports, scanned books, \emph{etc.}). Our system provides the facility to upload PDF or image documents and receive OCR output for the input document. We use Tesseract 4 to perform OCR, which is an LSTM-based \cite{greff2016lstm} \cite{saluja2019ocr} OCR engine. We have fine-tuned the model based on preliminary data collected from the tool on \textit{ Devanagari} font. Our fine-tuned model performs significantly better than the default version provided by Tesseract\footnote{https://tinyurl.com/san-tesseract}.

We use the Samanantar corpus \cite{ramesh2021samanantar} consisting of 8.46 million parallel sentences in English and Hindi to train a bilingual English-Hindi model. We build our NMT model based on Transformer \cite{vaswani2017attention} using fairseq \cite{ott2019fairseq}. Byte-pair encoding is used to develop the subword vocabulary. Each model contains 6 encoder layers, 6 decoder layers and 8 attention heads per layer. The input embeddings are of size 512, the output of each attention head is of size 64 and the feedforward layer in the transformer block has 2048 neurons \cite{ramesh2021samanantar}.
\subsubsection{Lexically constrained translation}
NMT models are biased towards the training corpus and produce domain-oblivious translations. Lexically constrained translation allows ingestion of domain-specific terminology and other phrasal constraints during decoding \cite{hokamp-liu-2017-lexically, alkhouli-etal-2018-alignment}. Current tools do not ingest lexical constraints while producing initial machine translation. We use alignment-based constraint decoding methods that ingest domain-specific terminology without requiring training domain-specific NMT models. These alignment-based constrained methods \cite{chen2021lexically,hokamp-liu-2017-lexically} captures source to target alignment using weights of intermediate decoder and encoder-decoder cross attention layers. We can determine whether a source constraint is currently being translated and then revise target tokens based on the corresponding target constraints. Inspired by Leca \cite{chen2021lexically}, we train multi-lingual transformer-based NMT models using Samanantar dataset \cite{maheshwari2022dictdis} . We adopt this method to produce domain-aware raw MT, thus reducing the effort required for post-editing, as reported in our user study (Section \ref{sec:userstudy}).

\begin{figure}[!t]
    \centering
    \includegraphics[width=0.99\linewidth]{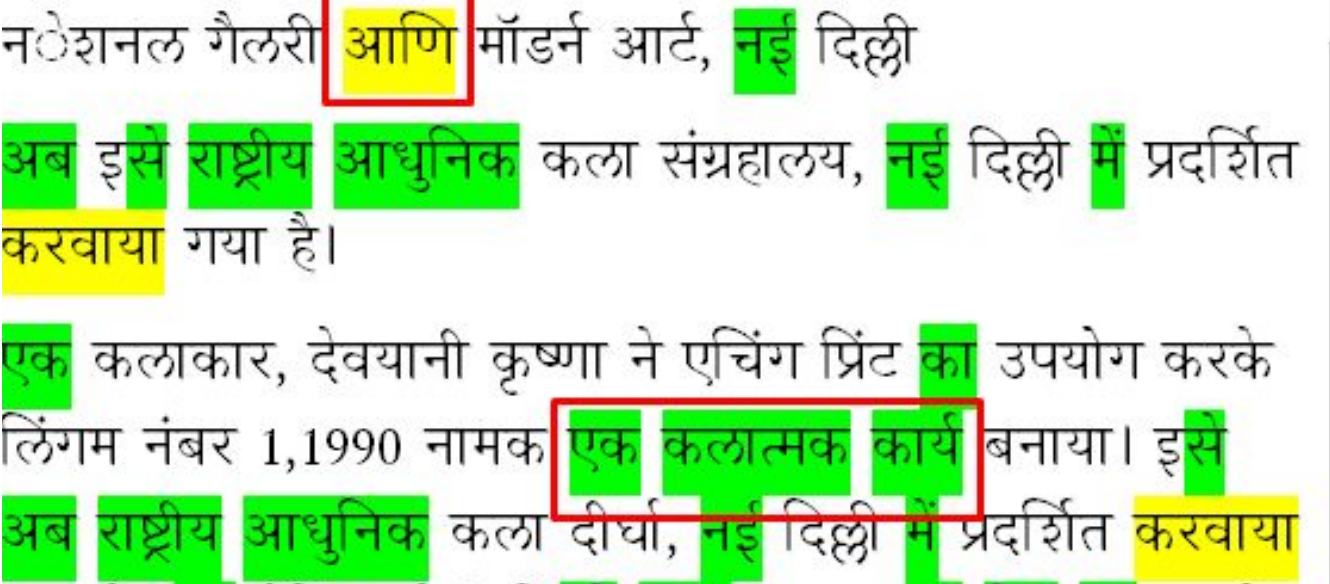}
    \caption{Users can track edits based on global replacements or technical dictionary suggestions implemented by highlighting the words using different colours. Yellow highlight refers to globally replaced words and green highlight refers to dictionary word replacement.}
    \label{fig:color-replace}
\end{figure}

\subsection{Word Alignment}
Existing CAT tools \cite{alabau-etal-2014-casmacat, lee2021intellicat} align source and target words and highlight their correspondence when the user places the mouse or text cursor on a word. However, these tools accept sentence as an input while our tool ingests PDF or image documents as an input. Hence, we obtain the page-wise OCR of the document and word-level alignments. The alignments are obtained by fine-tuning multilingual-BERT embeddings using AwesomeAlign \cite{awesome-align}. AwesomeAlign is an unsupervised approach to increase alignment between words having similar representation and reduce spurious alignments. It uses a combination of mask language modeling and self-training objective to encourage similarity between source and target by exploiting their agreement. We use a Samanantar dataset \cite{ramesh2021samanantar} to fine-tune AwesomeAlign. 
\\
\textbf{Inference} AwesomeAlign deduces the final alignment between words by taking the intersection of source-to-target and target-to-source matrices, assuming a defined threshold. We use a greedy algorithm, similar to simAlign \cite{sabet2020simalign}, for inference. We observed that lesser words are dropped, leading to better alignment. 

\subsection{TM and Domain specific lexicon}
The tool provides several ways to include translation memory from the source-target, target-target and domain-specific lexicons. \\
 \textbf{Translation Memory (TM)}: Existing tools \cite{federico2014matecat, lee2021intellicat, alabau-etal-2014-casmacat} provide the facility of TM to assist post-editing. These tools perform the sentence-wise translation in the interface where the source and target sentences are available in a parse-able format, thus making it easier to store source and target translations. In contrast, our tool performs document-level translation wherein it is difficult to keep track of the corresponding source and target sentences. Instead, we use word alignment to induce source-target translation during pre-processing. \\
    \noindent \textbf{Target-Target PE}: We store target-target segment edits made by the users and allow users to download the segment edits. The user can in turn, upload the segment edits in the tool for a different document and replace the segments in a single click.\\
    \textbf{Domain specific lexicon}: In technical translation, a domain-specific lexicon helps to disambiguate translated words. Users can choose to include more than 100 lexicons from domains such as Physics, Mathematics, Chemistry, Computer Science, Banking, {\em etc.}  For each page, words from lexicons are uniquely identified using green color highlight (refer Figure \ref{fig:tool-image}).

\subsection{Global replacement of edit segments}
As shown in Figure \ref{fig:replacement}, when a user modifies and saves the target page, it is prompted with the list of edits made on the current page. The prompt provides an option to replace the edited words across all pages. The user can choose i. to replace words within the current page or ii. replace them in the unedited pages or iii. replace them globally across all pages. A preview option is provided to view the changes and aid the users in informed decisions on the global replacement strategy. To keep track of edits made using external lexicon and global replacements, they are highlighted in different colours to aid the editing experience, as shown in Figure \ref{fig:color-replace}.

\subsection{Logging \& Multi-user access}

The tool logs following unit of information during a PE session:. i. Number of edits a user makes on a page, ii. time spent on each page. iii. Word replacement (either global or single replacement) and the word that was being replaced.
The tool is designed such that multiple post-editors can work on the same project. The tool saves the project in the user's local system and pushes the project to a remote git repository (in our case, Github). Thus, multiple post-editors can work on different document pages simultaneously.

\section{User Study}\label{sec:userstudy}
We conducted a user study to evaluate the effectiveness of the UDAAN tool. We measured the time spent on each page during a performance evaluation session. We compared this with two other methods: Translating from scratch and post-editing the MT output in a simple word editor.
\noindent \textbf{Participants and Study Design}: We recruited 6 participants (aged 25-50 years) from our user-base working in different domains such as banks, publications houses, \emph{etc.,} (for a detailed list refer \href{https://udaanproject.org}{project website}), who are fluent in both English and Hindi. The participants were familiar with the tool, having used it to translate one or more documents. They were asked to translate English documents into Hindi.

\noindent \textbf{Translation Interfaces} We use three translation interfaces to test our system: \textbf{MS-Word}, \textbf{MT-Only} and \textbf{UDAAN}. For MS-Word, given a PDF document, the user was asked to type the translated text from scratch in a word document. In MT-only, participants are provided with the MT document and its corresponding PDF and are asked to edit them in a word document. In the UDAAN setup, we ask participants to use our tool for PE. The English documents consist of 3 pages each and are from three domains: finance, mechanics, and literature consisting of total 1169, 1339 and 747 words respectively. 


In Table \ref{tab:user-study-results}, we summarise the results of our user study. As shown, our tool speeds up the translation by approximately a factor of three. Additionally, the standard deviation measured across different participants is lower with our tool. 
The translators produce similar translations of the given documents without much deviation in semantics of the translated sentences. This establishes the fact that the performance is consistent across users.


\begin{table}[!h]
\centering
\begin{tabular}{@{}lccc@{}}
\toprule
Document & MS-Word & MT-Only & UDAAN Tool \\ \midrule
Finance & 78.3 ({\small{25.7}}) & 70.3({\small{51.3}}) & \textbf{27.7({\small{15}})} \\
Technical & 94.7({\small{77.4}}) & 52.3 ({\small{62.9}})& \textbf{27.0({\small{5.2}})} \\
Literature & 88.3 ({\small{52.6}})& 68.3({\small{66.4}}) & \textbf{29.0({\small{9.6}})} \\ \bottomrule
\end{tabular}
\caption{Average time to complete translation (in minutes). Numbers in brackets `()' represent the standard deviation of translation time.}
\label{tab:user-study-results}
\end{table}

\section{Conclusion and Future Work}
We introduce UDAAN, a post-editing tool for translation, and OCR. The pipeline of UDAAN consists of producing lexically constrained MT, which is further used for post-editing using the tool. It provides several post-editing features that significantly improve the quality of translation. Compared to translation from scratch and post-editing over MT, the user study shows that our tool achieves 3-3.5 times increase in speed. Finally, the applicability of our tool was well evaluated in the user study. We plan to develop a web interface that will allow users to work in an online mode.
In our upcoming work, we plan to make our replacement module robust by including context-sensitive and morphologically-aware replacement techniques for suggesting replacements. Based on the candidate replacement patterns generated during post-editing stage, we build upon \cite{abhishek2021spear, maheshwari2022learning} to produce syntactically and semantically similar patterns for replacements. This will help reduce the post-editing effort.  We plan to build a pipeline that improves our MT performance continuously from the user-edited triplet of source segment, MT segment, and post-edited segment. Finally, we will develop a web interface that will allow users to work in an online mode.
\begin{acks}
Ayush Maheshwari is supported by a Fellowship from Ekal Foundation (www.ekal.org). Ganesh Ramakrishnan is grateful to NLTM OCR Bhashini project as well as the IIT Bombay Institute Chair Professorship for
their support and sponsorship. We thank Anuja Dumada, Pranita Harpale and Akshay Jalan for their support in the tool development and management.
\end{acks}
\balance
\bibliographystyle{ACM-Reference-Format}
\bibliography{ref}




\end{document}